\documentclass[conference]{IEEEtran}
\IEEEoverridecommandlockouts
\usepackage{cite}
\usepackage{amsmath,amssymb,amsfonts}
\usepackage{algorithmic}
\usepackage{graphicx}
\usepackage{textcomp}
\usepackage{xcolor}
\usepackage{enumitem}
\usepackage{comment}
\usepackage{booktabs}

\usepackage{subfig}
\usepackage{dblfloatfix}
\usepackage{amsfonts}
\usepackage{mwe}

\usepackage{scalerel}

\newcommand{\mat}[1]{\mathbf{#1}}

\DeclareMathOperator{\E}{\mathbb{E}}

\begin{document}

\title{Guided-GAN: Adversarial Representation Learning for Activity Recognition with Wearables}

\author{\IEEEauthorblockN{Alireza Abedin\IEEEauthorrefmark{1},
Hamid Rezatofighi\IEEEauthorrefmark{2} and
Damith C. Ranasinghe\IEEEauthorrefmark{1}
}
\IEEEauthorblockA{\textit{\IEEEauthorrefmark{1}The School of Computer Science, 
The University of Adelaide}, SA 5005 Australia\\
\{alireza.abedinvaramin, damith.ranasinghe\}@adelaide.edu.au }
\IEEEauthorblockA{\textit{\IEEEauthorrefmark{2}Monash University}, 
Melbourne, VIC 3800, Australia\\ 
hamid.rezatofighi@monash.edu}}

\maketitle

\begin{abstract}
	Human activity recognition (HAR) is an important research field in ubiquitous computing where the acquisition of large-scale labeled sensor data is tedious, labor-intensive and time consuming.  State-of-the-art unsupervised remedies investigated to alleviate the burdens of data annotations in HAR mainly explore training autoencoder frameworks. 
	In this paper: we explore generative adversarial network (GAN) paradigms to \textit{learn unsupervised feature representations from wearable sensor data}; and design a new GAN framework---\underline{G}eometrically-G\underline{uided} GAN or \textit{Guided-GAN}---for the task. To demonstrate the effectiveness of our formulation, we evaluate the features learned by Guided-GAN in an \textit{unsupervised} manner on three downstream classification benchmarks. Our results demonstrate Guided-GAN to outperform existing unsupervised approaches whilst closely approaching the performance with fully supervised learned representations. The proposed approach paves the way to bridge the gap between unsupervised and supervised human activity recognition whilst helping to reduce the cost of human data annotation tasks. 
\end{abstract}


\section{Introduction}

The proliferation of low-cost sensing technologies coupled with rapid advances  in machine learning techniques are enabling automatic human activity recognition (HAR) using wearables to realise a multitude of applications, especially in the healthcare domain~\cite{mannini2017activity,robertoFallsPO2017,Govercin2010UserReqFallDetect,healthapp,pd,asangiweariswc,health4,chesser2018bedexit}. However, the growth and performance of wearable based activity recognition applications are impeded by the demand for annotated sensor data collection, a process both expensive and laborious to undertake, for supervised learning methods. Consequently, investigating unsupervised HAR to avoid the heavy reliance on labeled data, has sparked significant interest. But, despite advances in unsupervised learning methods, investigations of unsupervised representation learning for HAR applications using wearable sensor data remains little explored.

The state-of-the-art methods for unsupervised representation learning have largely explored autoencoder frameworks~\cite{role,motion2vec}; an encoder first projects the data into a compact latent representation and a decoder exploits the representation to sequentially re-generate the original sensor data. Within this framework, the network weights are encouraged to learn feature representations that minimize the element-wise reconstruction error. However, it is not clear whether the \textit{pretext} task of element-wise reconstruction alone suffices for extracting enriched activity features. In contrast, in the vision domain, exploration of the latent spaces in generative adversarial networks (GANs) \cite{gan} built upon deep convolutional neural networks has resulted in promising frameworks for unsupervised learning of enriched feature representations~\cite{advfeature,ali,icgan,infogan,faae,bigbigan}. 

\vspace{1mm}
\noindent\textbf{Our Motivations.~}Despite the demonstrated strength of GANs in the visual domain to capture semantic variations of data distributions, the adoption of GANs for the challenging task of unsupervised representation learning for sequential wearable data remains. In contrast to vision tasks, multimodal sequential data, typical of those from wearables, are uniquely characterized by the inherent sample dependencies across time and desire architecture designs beyond convolutional operators for temporal modeling. But, examination of GANs for temporal HAR data are predominantly confined to synthesizing artificial sequences that resemble the original data \cite{crnngan,medical,sensegen,sensorygans,wifi,timegan}, while investigation of GAN's latent feature space for unsupervised learning remains largely unexplored. Further, GANs are notorious for their unstable training process and sensitivity to hyper-parameter selections. Although, the immense community effort in computer vision has resulted in established guidelines for architectural designs---weight initializations and hyper-parameter settings~\cite{dcgan,miyato,salimans,karras}---the same exploration is lacking in the sensor-based HAR domain. \textit{Motivated by these factors and grounded on the immense success of GANs in the visual domain, we believe exploring the GAN's latent feature space offers an appealing alternative to the de-facto autoencoder-based frameworks \cite{audeep,motion2vec,role} for wearable sensor representation learning}.  

\begin{figure*}[t]
	\centering
	\subfloat{\includegraphics[width=0.85\textwidth, trim = {6mm 5mm 1mm 9mm}, clip]{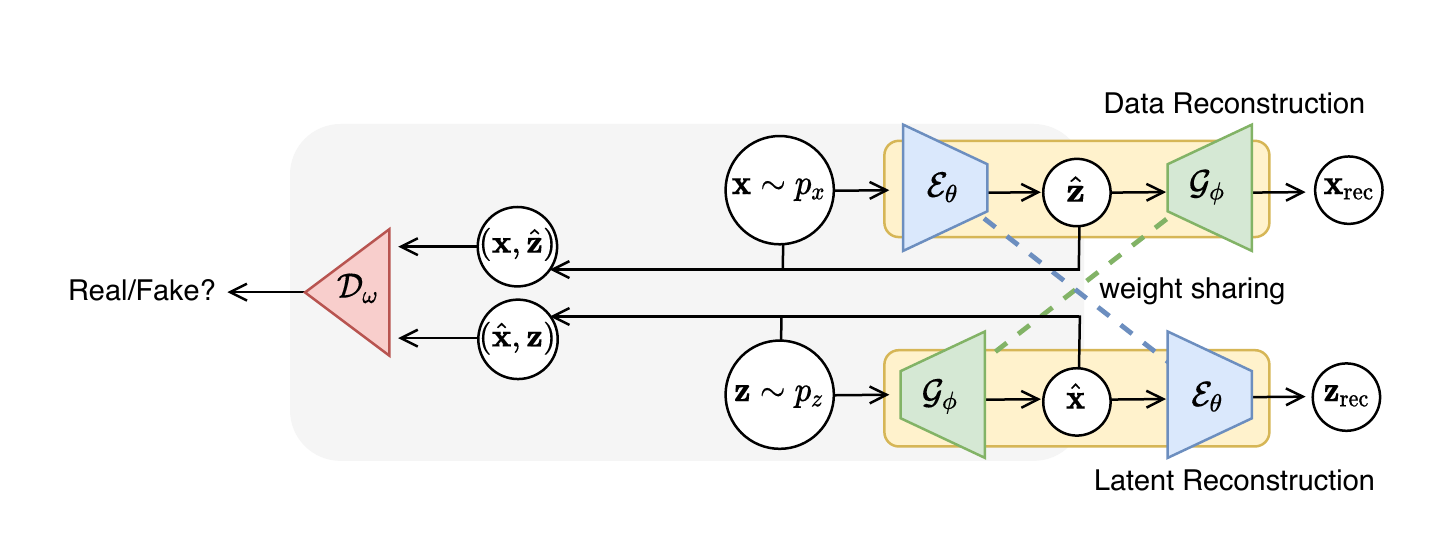}}%
	\caption{\textbf{Guided-GAN.} Overview of our proposed adversarial game for unsupervised representation learning from sequential wearable data. In relation to~\protect\cite{advfeature} (gray box), we propose: \textit{\textbf{i)}}~incorporating gradient feedback from geometric distance minimization in both data and latent manifolds; \textit{\textbf{ii)}}~efficient implementation architecture through parameter sharing (dashed lines); and \textit{\textbf{iii)}}~integration of our recurrent block designs in Fig.~\ref{gen_fig:ours_mod} for temporal modeling.}\vspace{-0.7em}
	\label{gen_fig:ours_arch}
\end{figure*}

\begin{figure*}[b]%
\vspace{-1.2em}
	\centering
	\subfloat{\includegraphics[width=0.8\textwidth]{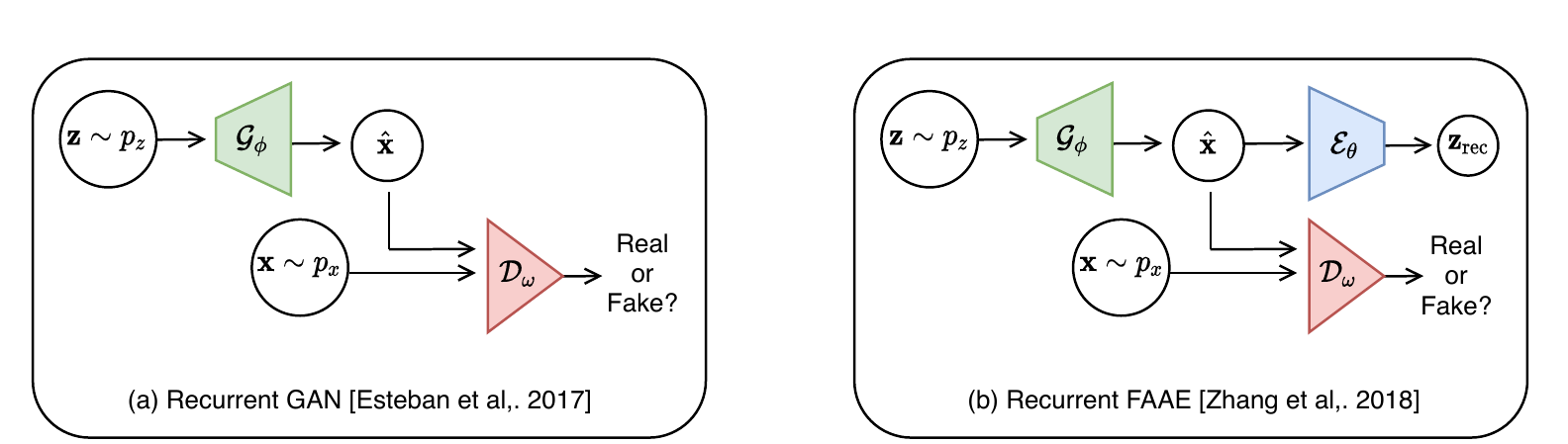}}
	\caption{Generative framework pipelines for (a) RGAN \protect\cite{medical}, and (b) Recurrent adaptation of flipped adversarial autoencoder proposed in \protect\cite{faae}. }
	\label{gen_fig:base_arch}\vspace{-1.2em}
\end{figure*}

\begin{figure*}[t]%
	\centering
	\subfloat{\includegraphics[width=0.9\textwidth]{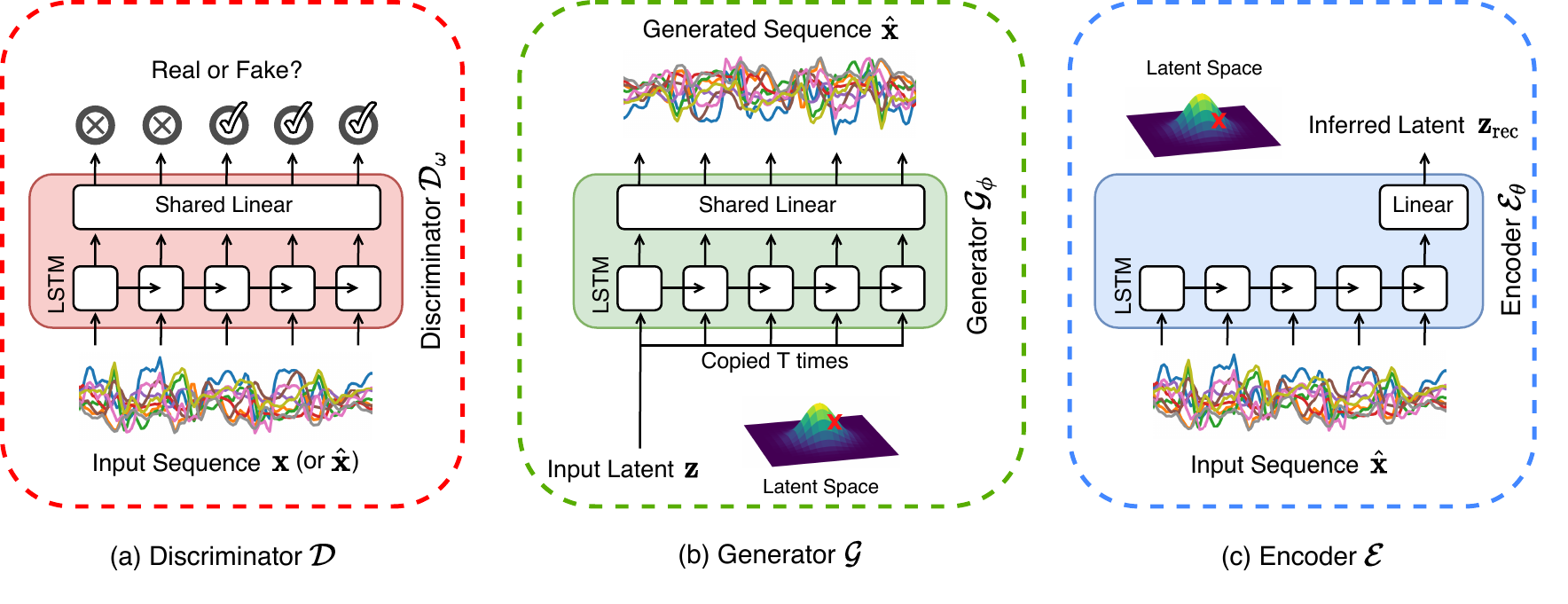}}
	\caption{An illustration of baselines' recurrent building blocks: (a) Discriminator functions in the data space to produce real vs. fake classification scores at each time-step. (b) Generator consumes a latent input repeated to the sequence length and produces a synthetic sample in the sequential data space. (c) Encoder serves as the inference machine and projects an input sequence into its corresponding latent representation. Notably, we depict a slightly modified version of \cite{medical}; \textit{i.e.}: i) a shared linear layer is added on top of the recurrent networks; and ii) instead of sampling independent latents, a single latent is sampled and replicated to the sequence length.}
	\label{gen_fig:base_mod}\vspace{-1.2em}
\end{figure*}

\noindent\textbf{Our Approach.~}We design a generative adversarial framework for unsupervised representation learning from wearables. Our intuitive approach measures and minimises the errors associated with reconstructing \textit{data and latent samples} and is efficiently implemented through re-using existing components with \textit{weight sharing} as illustrated in Figure~\ref{gen_fig:ours_arch}. In summary:

\begin{itemize}[leftmargin=4mm]
    \item We present a first rigorous study of generative adversarial frameworks for unsupervised representation learning from sequential multi-modal data from wearable sensors.
    \item We develop a \textit{novel} unsupervised representation learning framework with a generative adversarial network architecture. Our framework augments adversarial feedback with geometric distance guidance to encourage the encoder to invert the generator mappings with symmetrically orchestrated recurrent generator and encoder components.  Exploiting the symmetry, we craft \textit{an efficient implementation architecture} with weight sharing. 
    \item We conduct a series of systematic experiments to demonstrate the effectiveness of our proposed approach for unsupervised representation learning from sequential multi-modal data through downstream classification tasks.
\end{itemize}

\section{Background}\label{sec:backg}
Multi-modal sensing platforms continuously record measurements through different sensor channels over time and generate sequential multi-modal data. The acquired stream is then partitioned into segments $\mat{x}\in\mathbb{R}^{\textrm{D}\times \textrm{W}}$ using a sliding window, where $\textrm{D}$ denotes the number of sensing modalities used for data acquisition and $\textrm{W}$ represents the choice for the window duration. Here, the goal is to learn unsupervised representations enriched with distinctive features that can subsequently benefit classification of generated sequences. In what follows, we first discuss how existing GAN frameworks can be adopted for unsupervised representation learning of such sequences (Section~\ref{sec:backg}). Highlighting the existing challenges, we then introduce our novel framework to uncover unsupervised representations with higher correspondence to class semantics (Section~\ref{sec:our-framework}).

\subsection{Recurrent Generative Adversarial Networks}\label{sec:rgan}

The standard GAN~\cite{gan} comprises of two parameterized feed-forward neural networks---a generator $\mathcal{G}_{\phi}$ and a discriminator $\mathcal{D}_{\omega}$---competing against one another in a minimax game. Ultimately, the goal is for the generator to capture the underlying data distribution $p_x$. To this end, the generator exploits a simple prior distribution $p_{z}$ to produce realistic samples that trick the discriminator. On the contrary, the discriminator is trained to distinguish between the real and the generated samples. The resulting adversarial game optimises 
\begin{equation}\label{eq:loss_gan}
\min_{\mathcal{G}_{\phi}} \max_{\mathcal{D}_{\omega}} 
\E_{\mat{x}\sim p_{x}}[\log\mathcal{D}_{\omega}(\mat{x})] + 
\E_{\mat{z}\sim p_{z}}[\log(1-\mathcal{D}_{\omega}(\mathcal{G}_{\phi}(\mat{z})))],
\end{equation}
where, $\mat{x}\sim p_x$ represent the mini-batch training samples and $\mat{z}\sim p_z$ denote the drawn latents.

Extending the vanilla GAN to generate sequences of real-valued data, \cite{medical} substitutes both the generator and discriminator with recurrent neural networks and develops the Recurrent GAN (RGAN) for medical time-series generation. Within the resulting framework depicted in Fig.~\ref{gen_fig:base_arch}-a, the generator $\mathcal{G}_{\phi}$ takes a latent sample $\mat{z}$ and sequentially generates multi-channel data for each time-step. Similarly, the discriminator $\mathcal{D}_{\omega}$ consumes an input sequence and delivers per time-step classification decisions. We visualize the internal structure of these components in Fig.~\ref{gen_fig:base_mod}.

While the focus in \cite{medical} is solely on sequence generation, in our experimental study, re-investigate the framework for the purpose of unsupervised feature learning for sequences; the intermediate representations from the trained discriminator of a GAN are found to capture useful feature representations for related supervised tasks \cite{dcgan}. Intuitively, these set of features are attained free of cost and encoded in the discriminator weights when distinguishing real sequences from generated sequences during training. Notably, RGAN provides the arguably most straightforward extension of a regular GAN for the sequential domain and thus, we base our own investigations by building upon this framework. 

\begin{figure*}[t]
	\vspace{-0.5em}
	\centering
	\includegraphics[width=1.02\textwidth]{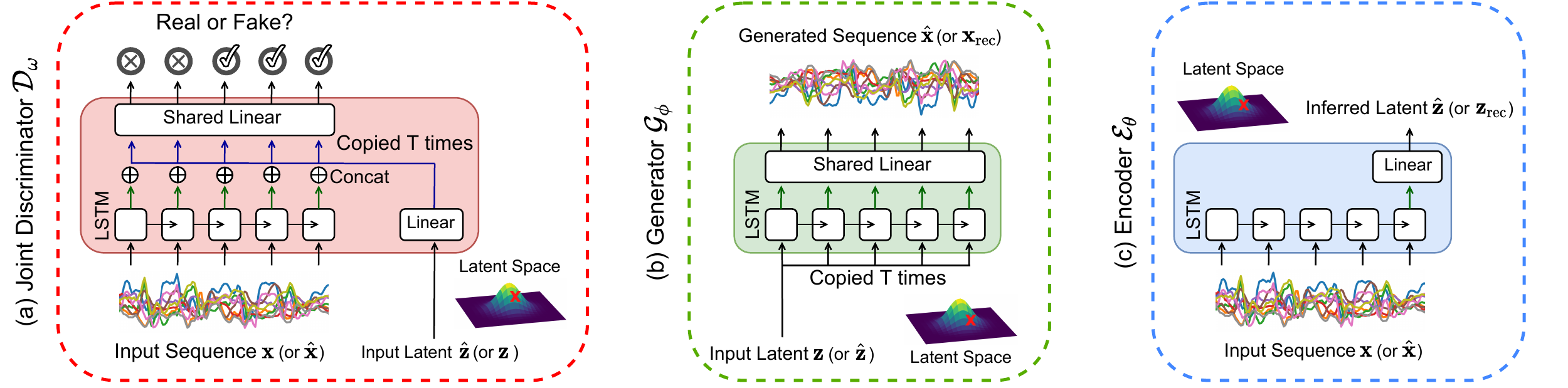}
	\caption{Our developed recurrent components: (a) Discriminator receives pairs of data-latents, learns a sequence of aggregated feature representations and delivers per time-step classification scores; (b) Generator and (c) Encoder leverage the same architecture design as RGAN and FAAE. But, they are trained to minimize the associated reconstruction errors in \textit{data and latent} spaces, respectively. $\bigoplus$ represents the concatenation operation of projected data and latent features.}\vspace{-1.2em}
	\label{gen_fig:ours_mod}%
\end{figure*}

\subsection{Recurrent Flipped Adversarial AutoEncoder}

Despite their empirical strength to model arbitrary data distributions, the vanilla GAN and in turn the RGAN, naturally lack an inference mechanism to directly infer the latent representation $\mat{z}$ for a given data sample $\mat{x}$. Accordingly, \cite{faae} proposes a natural extension of GANs to jointly train an encoder network that embodies the inverse mapping and coins the name Flipped Adversarial AutoEncoder (FAAE). Visualized in Fig.~\ref{gen_fig:base_arch}-b, the resulting framework exploits the adversarial guidance of a discriminator in the data space exactly identical to a regular GAN. In order to train the encoder, it additionally minimizes the reconstruction error associated with reproducing the latent representations. The training objective thus translates to:

\begin{equation}\label{eq:loss_faae}
\begin{split}
\min_{\mathcal{G}_{\phi},\mathcal{E}_{\theta}} \max_{\mathcal{D}_{\omega}} 
\E_{\mat{x}\sim p_{x}}[\log\mathcal{D}_{\omega}(\mat{x})] \\ &  + 
\E_{\mat{z}\sim p_{z}}[\log(1-\mathcal{D}_{\omega}(\mathcal{G}_{\phi}(\mat{z})))] \\ & +  \|\mat{z}-\underbrace{\mathcal{E}_{\theta}(\mathcal{G}_{\phi}(\mat{z}))}_{{\mat{z}_{\textrm{rec}}}}\|^2_2,
\end{split}
\end{equation}
where, $\mathcal{E}_{\theta}$ denotes the parameterized encoder network and  $\mat{z}_{\textrm{rec}}=\mathcal{E}_{\theta}(\mathcal{G}_{\phi}(\mat{z}))$ is the reconstructed latent representation. 

Since there exists no previous exploration of FAAE framework for the sequential domain, we design and augment RGAN with a recurrent encoder $\mathcal{E}_{\theta}$ depicted in Fig.~\ref{gen_fig:base_mod}. In particular, the encoder reads through the generated sequences from the generator and updates its internal hidden state according to the received measurements at each time step. Ultimately, the final hidden state after processing the entire sequence is exploited to regress the latent representations. In addition, the encoder and generator receive adversarial feedback from the discriminator for parameter updates during training. 

\subsection{Bi-directional Generative Adversarial Networks}

While the FAAE framework paves the way for learning the inverse mapping function, the encoder $\mathcal{E}_{\theta}$ performance is heavily reliant on the quality and diversity of generator's produced samples. Essentially, the encoder is never exposed to the original data from the training set and thus, its learned feature representations are handicapped by the generator's performance. Accordingly, \cite{advfeature,ali} propose the BiGAN with a novel approach to integrate efficient inference. We illustrate BiGAN while contrasting it with our proposed formulation, Guided-GAN, in the gray shaded area in Fig.~\ref{gen_fig:ours_arch}; \textit{i.e.}, the discriminator is modified to discriminate not only in the data space, but rather in the joint data-latent space between $(\mat{x}, \mathcal{E}_{\theta}(\mat{x}))$ and $(\mathcal{G}_{\phi}(\mat{z}), \mat{z})$ pairs. Hence, the corresponding minimax objective is defined as 
\begin{equation}\label{eq:loss_bigan}
\begin{split}
\min_{\mathcal{G}_{\phi},\mathcal{E}_{\theta}} \max_{\mathcal{D}_{\omega}} 
\E_{\mat{x}\sim p_{x}}[\log\mathcal{D}_{\omega}(\mat{x}, \mathcal{E}_{\theta}(\mat{x}))] \\ + \E_{\mat{z}\sim p_{z}}[\log(1-\mathcal{D}_{\omega}(\mathcal{G}_{\phi}(\mat{z}), \mat{z}))].
\end{split}
\end{equation}

To satisfy the objective, the generator is motivated to produce samples resembling the real data distribution and the encoder is incentivized to output latent representations matching with the prior latent distribution. It is shown in \cite{advfeature} that the theoretical optimal solution to this adversarial game leads to the encoder and generator inverting one another while the joint distributions are aligned. Importantly, the encoder in BiGAN has the luxury of directly learning from real samples $\mat{x}\sim p_x$.

\section{Our Proposed Framework}\label{sec:our-framework}

We formulate a novel framework, as illustrated in Fig.~\ref{gen_fig:ours_arch}, to uncover unsupervised representations with higher correspondence to class semantics by drawing inspiration from  BiGANs. We contrast our proposed formulation, Guided-GAN with the BiGAN components in Fig.~\ref{gen_fig:ours_arch}; specifically, our discriminator design not only discriminates in the data space, but in the \textit{joint data-latent space} between $(\mat{x}, \mathcal{E}_{\theta}(\mat{x}))$ and $(\mathcal{G}_{\phi}(\mat{z}), \mat{z})$ pairs. Our design: \textbf{\textit{i})}~realises a state-of-the art GAN for representation learning from sequential multi-modal data; and \textbf{\textit{ii)}}~alleviates the convergence problem~\cite{ali,faae}. We explain our design and the efficient implementation architecture next.

\vspace{1mm}
\noindent\textbf{Geometrically-Guided Adversarial Feedback.~}We observe, that the generator and encoder within the BiGAN framework do not directly communicate. Thus, the discriminator alone bears the burden of matching the joint data-latent distributions and guiding the encoder and generator components towards inverting one another at the optimal solution. Unfortunately, converging to the optimal theoretical solution is difficult to achieve in practice; thus, the encoder and the generator do not necessarily invert one another~\cite{ali,faae}. Hence, in addition to the adversarial feedback provided by the discriminator to match the joint data-latent distribution, we optimize geometric distance functions to match the marginal manifolds independently; \textit{i.e.}, we receive gradients from aligning: \textbf{\textbf{i)}}~the original data manifold with generator's induced output manifold; and \textbf{\textbf{ii)}}~the prior latent manifold with the encoder's output manifold. In particular for early stages of training, geometric distance optimization usually provides much easier training gradients~\cite{modegan}. \textit{We discovered this to be a vital necessity for successful training of a BiGAN in the sequential domain where GAN heuristics may be missing}. Notably, our attempts to train recurrent BiGANs without our proposed manifold distance minimization terms were unsuccessful---see Appendix~\ref{apd:biganexp}. Specifically, the encoder did not learn useful representations and resulted in extremely low downstream classification performance. To this end, we measure and minimize the reconstruction errors associated with reproducing both the data and the latent representations.  Hence, our proposed minimax adversarial game is formulated as: 
\begin{equation*}\label{eq:loss_ours}
\begin{split}
\min_{\mathcal{G}_{\phi},\mathcal{E}_{\theta}} \max_{\mathcal{D}_{\omega}} 
& \E_{\mat{x}\sim p_{x}}[\log\mathcal{D}_{\omega}(\mat{x}, \mathcal{E}_{\theta}(\mat{x}))] \\ & + 
\E_{\mat{z}\sim p_{z}}[\log(1-\mathcal{D}_{\omega}(\mathcal{G}_{\phi}(\mat{z}), \mat{z}))] \\ & + \lambda_x\|\mat{x}-\underbrace{\mathcal{G}_{\phi}(\mathcal{E}_{\theta}(\mat{x}))}_{{\mat{x}_{\textrm{rec}}}}\|^2_2 + \lambda_z\|\mat{z}-\underbrace{\mathcal{E}_{\theta}(\mathcal{G}_{\phi}(\mat{z}))}_{{\mat{z}_{\textrm{rec}}}}\|^2_2,
\end{split}
\end{equation*}

where $\|.\|^2_2$ imposes the $\ell_2$ reconstruction in data and latent feature spaces. Notably, the reconstruction errors are efficiently computed at no extra model complexity cost through weight sharing depicted in Fig.~\ref{gen_fig:ours_arch}. Here, $\lambda_x$ and $\lambda_z$ denote the loss balance coefficients. 

\vspace{1mm}
\noindent\textbf{Recurrent Symmetrical Adversarial Framework.~} In order to exploit the sequential nature of multi-modal wearable data streams, the core building blocks of our framework are empowered by recurrent units with memory cells, as depicted in Fig.~\ref{gen_fig:ours_mod}. Moreover, the generator and encoder communicate in a \textit{symmetrical orchestration} to measure the manifold distances in data and latent spaces. 

\textit{\textbf{Recurrent Joint Discriminator}---}The discriminator $\mathcal{D}_{\omega}$ operates in the joint data-latent space attempting to differentiate joint input samples of $(\mat{x}, \mathcal{E}_{\theta}(\mat{x}))$ against $(\mathcal{G}_{\phi}(\mat{z}), \mat{z})$ pairs. Internally, the multi-modal stream is initially processed by an LSTM network yielding a sequence of hidden state representations. Simultaneously, the latent input is linearly projected and concatenated to the LSTM hidden states at each time-step. The resulting sequence aggregates the learned features from both data and latent spaces, and is used to construct per time-step classification decisions.

\textit{\textbf{Symmetrical Generator and Encoder}---}The generator $\mathcal{G}_{\phi}$ and encoder $\mathcal{E}_{\theta}$ of our framework 
are symmetrically connected and serve augmented responsibilities: \textit{\textbf{i)}}~unlike a recurrent GAN, the generator is additionally exposed to encoded latents from the encoder's posterior distribution ($\hat{\mat{z}}=\mathcal{E}_{\theta}(\mat{x}); \mat{x}\sim p_x$) and is trained to reconstruct the input sequence; and \textit{\textbf{ii)}}~the encoder observes the generator's outputs ($\hat{\mat{x}}=\mathcal{G}_{\phi}(\mat{z}); \mat{z}\sim p_z$) and learns to regress the corresponding latent representation. \textit{Notably, given the symmetrical architecture design, our encoder now has access to both the original data samples (in contrast to FAAE) as well as the newly generated data samples (as opposed to BiGAN) to uncover generalizable feature representations and serve as a fully fledged feature extractor.}

\section{Experimental Evaluations} \label{experiments}

To validate our framework, we employ HAR benchmark datasets UCI HAR \cite{uci} and USC-HAD \cite{usc} exhibiting diversity in terms of the sensing modalities used and activities. To \textbf{\textit{allow easy visual interpretation}} of the results, we further re-purpose the popular MNIST hand-written digits dataset \cite{mnist} and explore sequential digit classification, as in~\cite{medical}. We follow the standard evaluation protocol in~\cite{eval} to assess the quality of unsupervised representations achieved for downstream supervised classification tasks. We detail the fair evaluation protocol, parameter settings and input-data. 
\textit{In summary}, we train all network modules using the \textit{unlabeled} training split sequences. Subsequently, we freeze the \textit{feature extractor} parameters and leverage the \textit{training labels} to train a single linear classifier on the learned representations. Except for the RGAN baseline~\cite{medical}, the \textit{encoder} network $\mathcal{E}$ within the frameworks serves as the \textit{feature extractor}. In particular:

\begin{itemize}
    
\item We describe the adopted representation learning baselines ($\S$\ref{supp_base}) and the fair evaluation protocol following previous studies($\S$\ref{supp_eval}).
\item We evaluate the effectiveness of the learned representations by transferring them for use in downstream classification tasks ($\S$\ref{sup_down}).
\item We assess the generalization of the the feature representations learned by the investigated unsupervised approaches for \textit{each classification task} by training a single linear classifier on the \textit{frozen} learned representations($\S$\ref{sec:gen-capbility}).
\item To share further insights through extensive experiments; We assess the ability of the unconditional generator trained through Guided-GAN to produce diverse and realistic multi-modal sequences ($\S$\ref{sup_gen}) as well as evaluate and visualize the faithfulness of sequential multi-modal data reconstructions by our Guided-GAN ($\S$\ref{sup_faith}).
\end{itemize}

\begin{table*}[ht]
	\caption{Comparison of unsupervised learned representations when transferred for use on downstream classification tasks. All methods employ recurrent neural networks. (\textbf{$\ast$})~RFAAE is our recurrent adaptation of \protect\cite{faae}.}\label{gen_tab:classification}
	\centering
	\def\arraystretch{1.1}
\begin{tabular}{lcc|cc|cc}
\textbf{\textit{Recurrent Representation}} & \multicolumn{2}{c}{Seq. MNIST} & \multicolumn{2}{c}{UCI HAR} & \multicolumn{2}{c}{USC-HAD}\\
Learning Method                            & Acc.            & \multicolumn{1}{c}{f1-score} & Acc.            & \multicolumn{1}{c}{f1-score} & Acc.            & f1-score         \\ 
\toprule
RAND                                       & 51.4\%          & 48.8\%                       & 51.6\%          & 44.3\%                       & 34.3\%          & 21.2\%           \\
M2V                                        & 82.8\%          & 82.6\%                       & 70.9\%          & 69.8\%                       & 54.1\%          & 44.0\%           \\
RGAN                                       & 80.2\%          & 79.8\%                       & 76.3\%          & 75.6\%                       & 50.7\%          & 42.8\%           \\
RFAAE (See\textasciitilde{}$\ast$)         & 94.0\%          & 93.9\%                       & 88.5\%          & 88.4\%                       & 64.4\%          & 57.2\%           \\
RAE-$\ell_1$                               & 95.5\%          & 95.5\%                       & 87.1\%          & 87.1\%                       & 63.8\%          & 54.8\%           \\
RAE-$\ell_2$                               & 95.7\%          & 95.7\%                       & 87.2\%          & 87.2\%                       & 65.2\%          & 56.0\%           \\
\textbf{Ours~(Guided-GAN)}                 & \textbf{97.3\%} & \textbf{97.3\%}              & \textbf{89.0\%} & \textbf{88.9}                & \textbf{67.2\%} & \textbf{59.9\%}  \\ 
\toprule
SUP (\textit{fully supervised})            & 99.1\%          & 99.1\%                       & 91.1\%          & 91.0\%                       & 68.6\%          & 62.6\%          \\
\toprule
\end{tabular}
\end{table*}

\begin{figure*}[t]%
	\centering
	\subfloat{\includegraphics[width=0.34\textwidth]{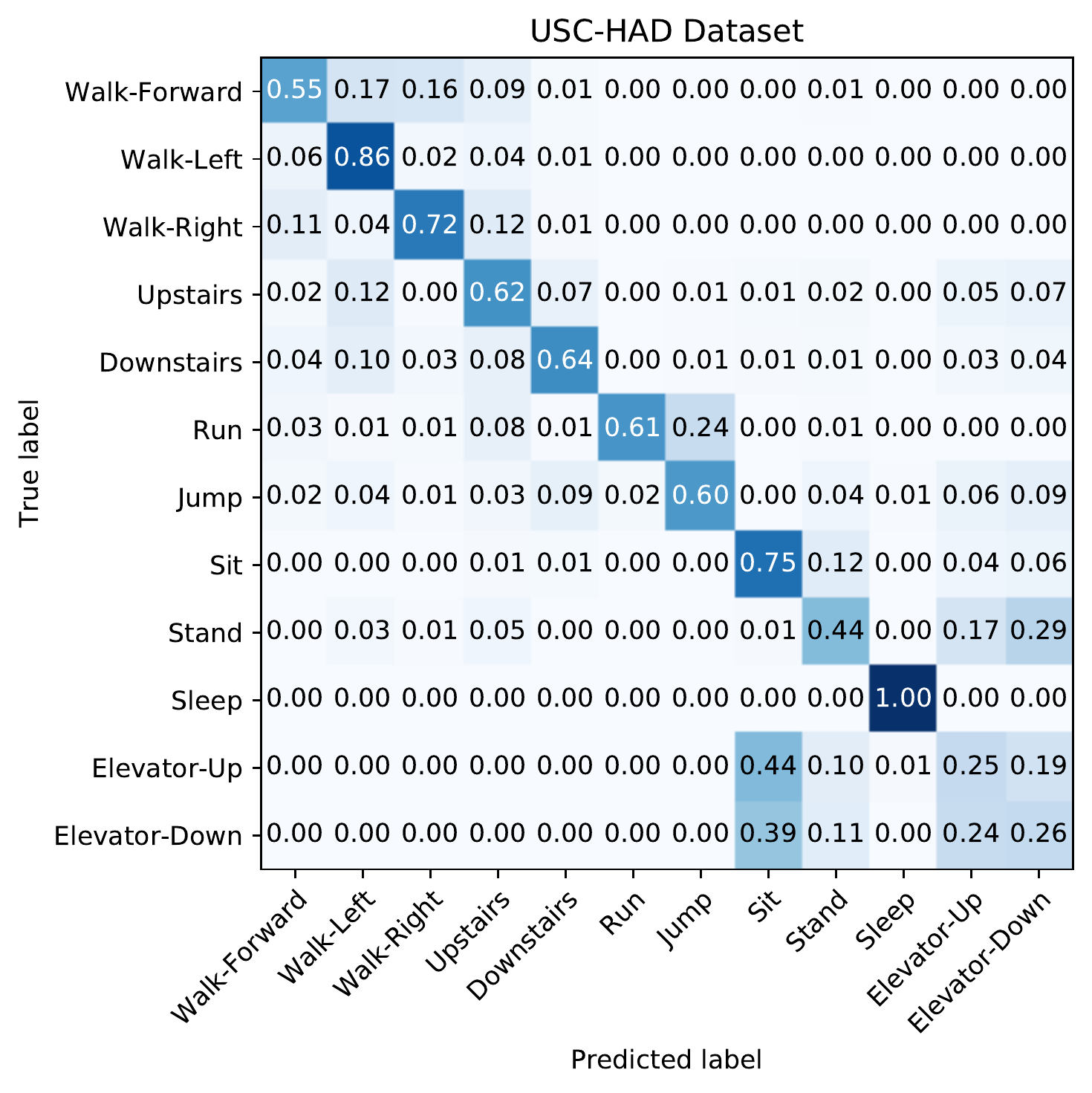}}
	\subfloat{\includegraphics[width=0.34\textwidth]{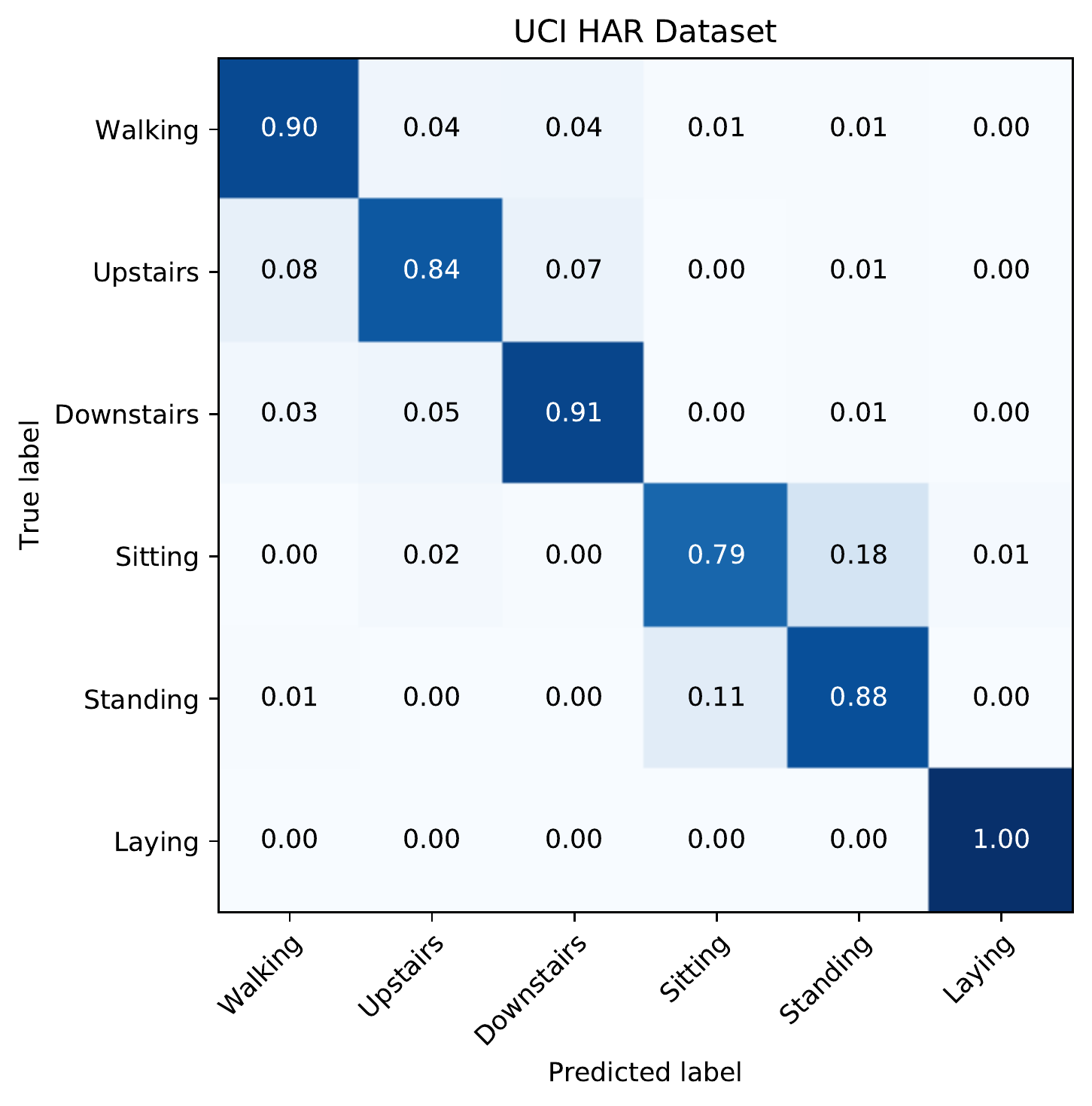}} 
	\subfloat{\includegraphics[width=0.33\textwidth]{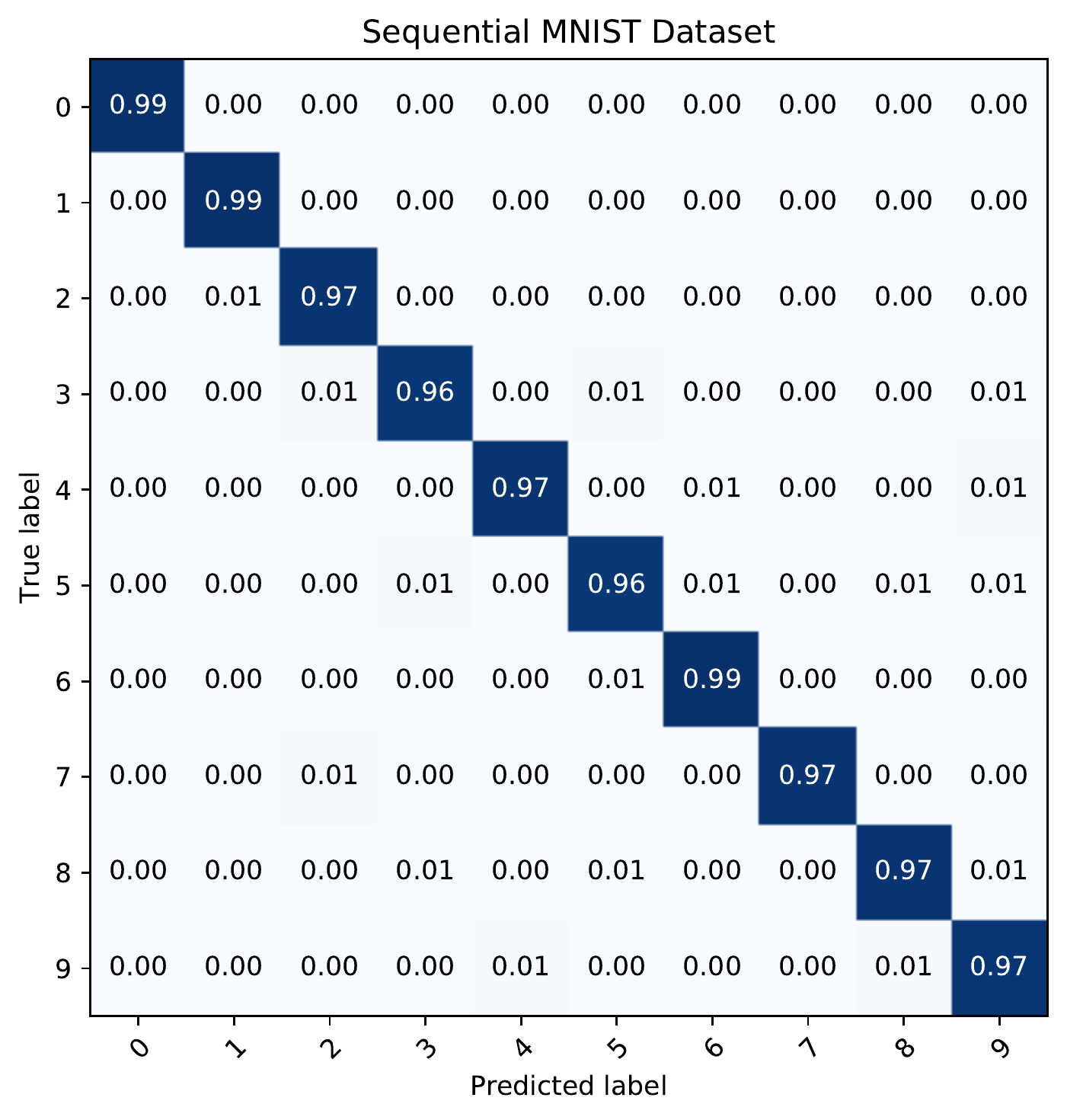}}
	\caption{Class-specific recognition results from the Guided-GAN's unsupervised features. The confusion matrices highlighting the class-specific recognition performance for the \textit{testing splits} of \texttt{Sequential MNIST}, \texttt{UCI HAR}, and \texttt{USC-HAD} benchmarks. The vertical axis represents the ground-truth labels and the horizontal axis denotes the predicted labels.}
	\label{confusion-mat}
\end{figure*}

\subsection{Unsupervised Activity Representation Learning Baselines}\label{supp_base}
We briefly introduce the alternative approaches that serve as concrete baselines for the task of unsupervised activity representation learning for HAR applications: 

\vspace{1mm}
\noindent\textit{Recurrent Autoencoder (RAE) \protect\cite{role}.~} The framework comprises of a deterministic encoder $\mathcal{E}$ and a decoder $\mathcal{G}$ trained directly to minimize $\ell_1$ or $\ell_2$ element-wise reconstruction error in the \textit{data space}.

\vspace{1mm}
\noindent\textit{Motion2Vector (M2V) \protect\cite{motion2vec}.~}
This framework includes a decoder $\mathcal{G}$, however with a stochastic encoder $\mathcal{E}$ parameterizing an Isotropic Gaussian $\mathcal{N}(\mathbf{0},\mathbf{I})$. The framework is trained with reconstruction error in the \textit{data space} as well as KL-divergence additionally optimized for the \textit{latent space}; KL-divergence is incorporate to match the encoder output distribution with the standard Gaussian prior. 

\vspace{1mm}
\noindent\textit{Recurrent Generative Adversarial Network (RGAN) \protect\cite{medical}.~}
The framework trains a generator $\mathcal{G}$ and a discriminator $\mathcal{D}$ by optimizing the standard adversarial loss in the \textit{data space} according to Eq. (1). in main manuscript. 

\vspace{1mm}
\noindent\textit{Recurrent Flipped Adversarial Autoencoder (RFAAE).~}We adapt the approach proposed in~\cite{faae} to a recurrent framework. In addition to the generator $\mathcal{G}$ and discriminator $\mathcal{D}$ of a standard GAN, this baseline jointly trains an additional encoder $\mathcal{E}$ to regress the latents; according to Eq.~(2), the adversarial loss is optimized in the \textit{data space}, and the $\ell_2$ reconstruction error is minimized in the \textit{latent space} between the encoder outputs and the sampled priors.

\subsection{Evaluation Protocol} \label{supp_eval}
Following~\cite{role}, sensory data are down-sampled to 33~Hz, and per-channel normalization is adopted using the training data statistics to scale the values into the range $[-1,1]$. Subsequently, the data-streams are partitioned into segments using a sliding window of 30 samples (\textit{i.e.}, W=30). 

We also follow the standard evaluation protocol in~\cite{eval} to assess the quality of unsupervised learned sequence representations of different baselines. All network parameters are trained end-to-end for 500 epochs by back-propagating the gradients of the corresponding loss functions averaged over mini-batches of size 64 and using the Adam~\cite{adam} update rule. The learning rate for Adam is set to $\textrm{10}^{\textrm{-3}}$ and the beta values $\beta_1=0.5$, $\beta_2=0.999$ are used. The fixed prior distribution $p_z$ for deep generative models is set to be a 100-dimensional isotropic Gaussian $\mathcal{N}(\mathbf{0},\mathbf{I})$. To ensure a fair comparison, the encoder $\mathcal{E}$, generator (interchangeably decoder in autoencoder frameworks) $\mathcal{G}$, and discriminator $\mathcal{D}$ constitute a single-layer uni-directional LSTM with 100 hidden neurons to process the input sequences. For our Guided-GAN, the loss weighting coefficients $\lambda_{z}=1$ and $\lambda_{x}=0.01$ are kept constant across the sensory datasets.

First, we train all network modules using the unlabeled training split sequences. Subsequently, we freeze the \textit{feature extractor} parameters and leverage the training labels to train a single linear classifier on the learned representations. Except for the RGAN baseline~\cite{medical}, the \textit{encoder} network $\mathcal{E}$ within the frameworks serves as the \textit{feature extractor}. For the RGAN baseline lacking an encoder module, the penultimate representations from the discriminator network $\mathcal{D}$ are used as the unsupervised features. The trained classifier is then evaluated on the held-out test split sequences and we report the achieved classification performance.

\begin{figure*}[t]%

	\centering
	\subfloat{\includegraphics[width=1.0\textwidth]{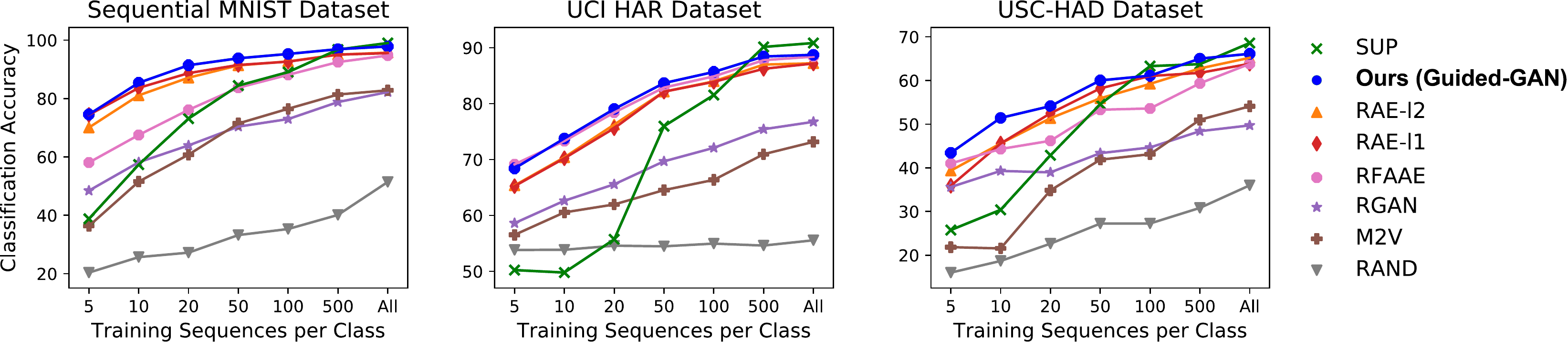}}%
	\caption{Effectiveness of learned sequence representations when building a classifier with varying sizes of labelled training data. We freeze the feature extractor parameters and employ training labels to train a single linear classifier on the learned representations. We train 1.6\%, 1.1\%, and 2.2\% of the parameters, respectively, on the three datasets and report mean values over the entire \textit{held-out test splits} in five runs with different subsets of training data.}
	\label{gen_fig:ratio}
\end{figure*}

\subsection{Downstream Sequence Classification}\label{sup_down} 
In Table \ref{gen_tab:classification}, we summarize the downstream classification performance by reporting the classification accuracy and class-averaged f1-score ($\textrm{F}_{m}$). Given the potential imbalanced class distributions, the latter metric reflects the ability of the HAR model to recognize every activity category regardless of its prevalence in the collected data. In Table \ref{gen_tab:classification}, RAE \cite{role} and M2V \cite{motion2vec} respectively employ recurrent autoencoder and variational autoencoder frameworks. We further present results from a \textit{fully supervised trained feature extractor} (\textbf{SUP}) and a randomly initialized feature extractor (\textbf{RAND}) for reference.

Across the three datasets, the large performance gap between RGAN (80.2\%, 76.3\% and 50.7\% respectively) against RFAAE (94\%, 88.5\% and 64.4\%) highlights the significance of incorporating an inference network for effective representation learning in generative adversarial frameworks. \textit{While the discriminator's delegated task of distinguishing between real and generated sequences benefits the penultimate representations (see the superiority of RGAN over RAND baseline), learning an inverse mapping to the latent feature space through encoder results in significantly more effective features}. However, the encoder in RFAAE is only trained on synthetic sequences and never encounters real data samples. Accordingly, its performance as a feature extractor is heavily reliant on the quality and diversity of the generator's sequences. In contrast, the encoder in our framework, \textit{exposed to both real data sequences as well as generated ones evidently offers feature representations of higher quality}, achieving 3.51\%, 0.56\% and 4.35\% relative improvements, respectively, on Sequential \texttt{MNIST}, \texttt{UCI HAR} and \texttt{USC-HAD} datasets.

Comparing the lower performance levels of M2V against its non-variational counterparts (RAE-$\ell_1$ and RAE-$\ell_2$), we observe that its ability to sample new sequences comes at the cost of harming the feature representation qualities. However, our proposed framework bridges this shortcoming by allowing the generation of realistic synthetic data while simultaneously achieving higher quality representations. \textit{Notably, our approach not only outperforms existing unsupervised baselines with a large margin but it also closely approaches the fully supervised baseline--\textit{SUP}--performance}.

\begin{table}[t]
	\caption{A detailed description on the number of trainable and frozen network parameters employed for downstream classification evaluation on the three benchmarks.}\label{nips:param}
	\centering
	\footnotesize		
	\def\arraystretch{1}
	\begin{tabular}{c|c|c|c}
		Parameter Count &  Sequential MNIST & UCI HAR  & USC-HAD \\
\toprule
		Total        & 63110   & 55106 & 54512\\ 
		Frozen       & 62100   & 54500 & 53300\\
		Trainable    & 1010    & 606   & 1212 \\
		\toprule
		(Trainable Ratio) & (\textbf{1.6\%})  & (\textbf{1.1\%})  & (\textbf{2.2\%}) \\ 
		\toprule
	\end{tabular}%
\end{table}

We summarise the number of parameters corresponding to the feature extractor (\textit{frozen}), classifier (\textit{trainable}) and the ratio of trainable parameters to the total number of network parameters (\textit{trainable ratio}) in Table \ref{nips:param}. \textit{We can observe the proposed Guided-GAN to achieve comparable classification performance to that of the fully supervised baseline (SUP) whilst having access to only \textbf{1.6\%}, \textbf{1.1\%}, and \textbf{2.2\%} of the parameters for training on the three datasets, respectively.}

For reference, we summarise the class-specific recognition results from the Guided-GAN's unsupervised features by presenting confusion matrices for the downstream classification tasks in Fig.~\ref{confusion-mat}.

\subsection{Generalisation of the Feature Representations}\label{sec:gen-capbility}
To gain further insights into the generalization capability of feature representations learned by the investigated unsupervised approaches, we analyze the classification performance on the entire testing splits for the three datasets while \textit{varying the amount of available labeled data for supervised classifier training} in Fig. \ref{gen_fig:ratio}. The reported results are averaged over five runs with different subsets of training data. 

We observe the unsupervised baselines to provide an effective means to learn useful feature representations by exploiting unlabeled data in the absence of large amounts of annotated training data, resulting in substantial performance gains over the \textit{RAND} and \textit{SUP} baselines; supervised classifier training on top of a randomly initialized feature extractor (\textit{RAND} baseline) fails to learn clear classification boundaries to discriminate different classes regardless of the amount of available labeled data and the fully supervised baseline (\textit{SUP} baseline) struggles to generalize to the unseen test sequences when trained on low volumes of annotated samples. But: i) our approach consistently offers better generalization to unseen data compared with existing unsupervised remedies in the presence of extremely limited labeled training data; and ii) Guided-GAN achieves higher classification performance when leveraging all labeled training data and is comparable with the supervised baselines trained end-to-end with \textit{full} supervision. 

\begin{figure*}[h]%
	\centering
	\subfloat{\includegraphics[width=0.9\textwidth]{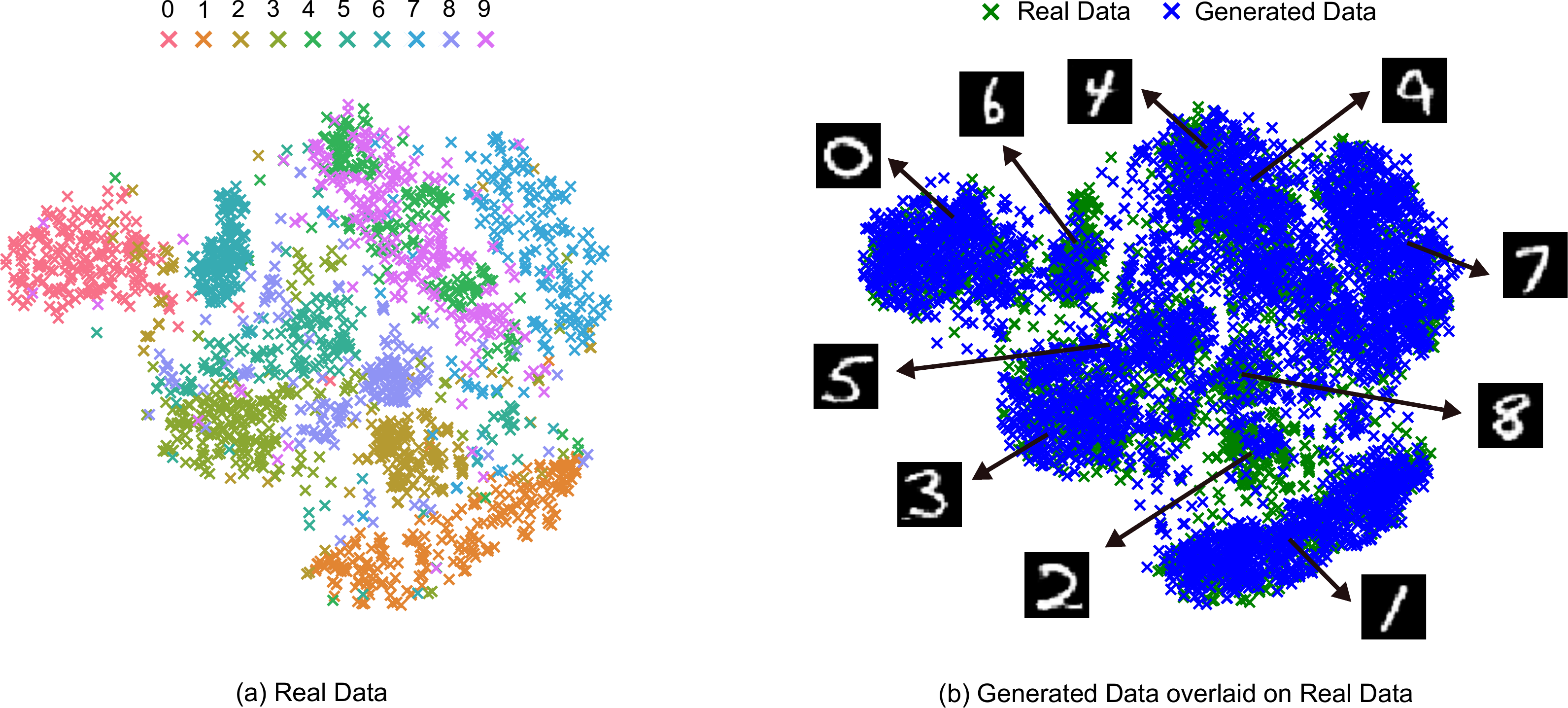}}%

	\caption{\textbf{\textit{Sequential model added for ease of visual interpretation}.} For \texttt{Sequential MNIST} dataset, we present t-SNE visualisation of: (a) real data sequences colour-coded with semantic labels, and (b) generated sequences overlaid on real data samples. Evidently, our Guided-GAN's generator successfully captures semantic variations in the data and aligns with the real data distribution.}
	\label{nips:mnist_space}

\end{figure*}
\begin{figure*}[t]%
	\centering
	\subfloat{\includegraphics[width=0.9\textwidth]{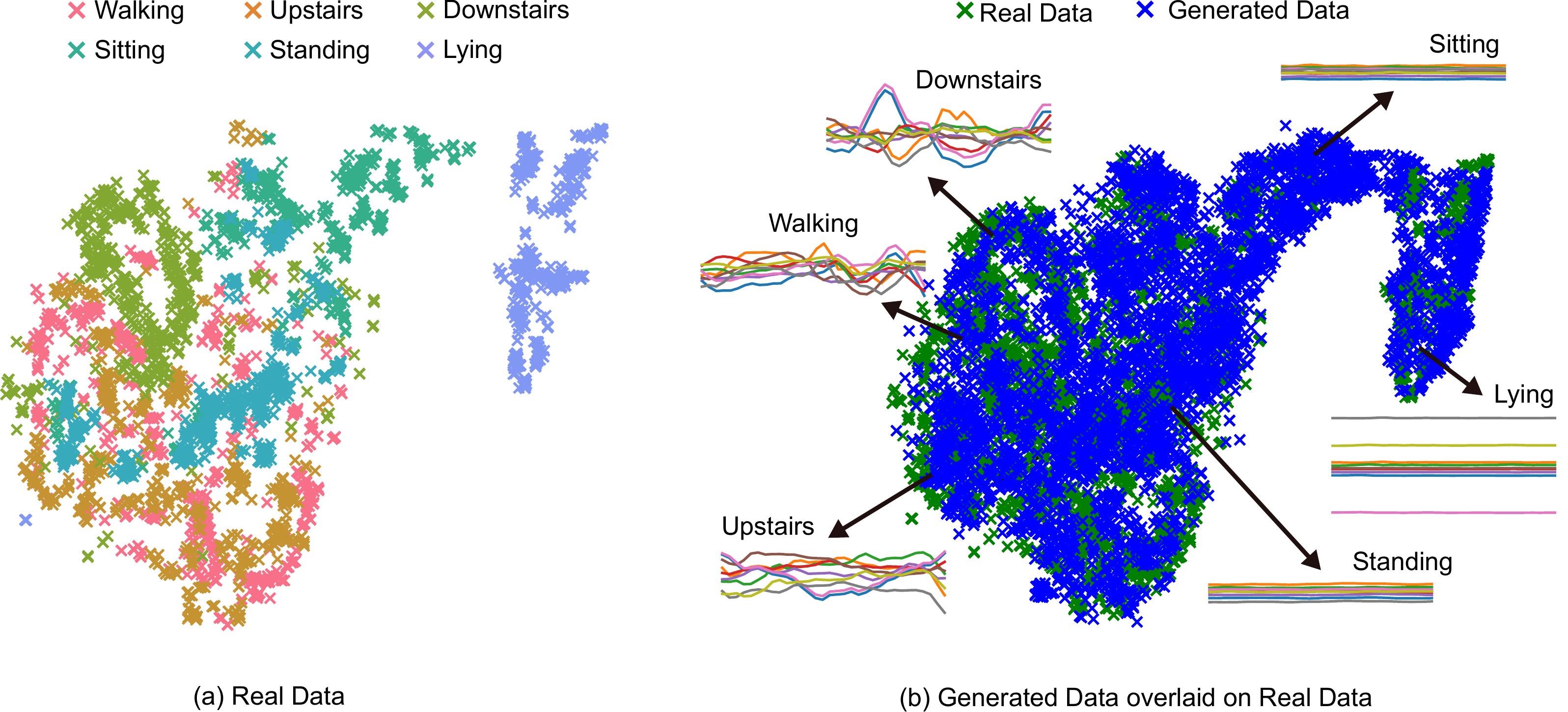}}%
	\caption{For \texttt{UCI HAR} benchmark, we present t-SNE visualisation of: (a) real data sequences colour-coded with semantic activity labels, and (b) generated sequences overlaid on real data samples.  Our Guided-GAN's generator is observed to successfully capture semantic variations embedded in the multi-modal motion sequences.}
	\label{nips:uci_space}
\end{figure*}

\subsection{Assessment of Multi-modal Sequence Generation}\label{sup_gen}
While \textit{unsupervised feature learning} constitutes the main focus of our study, we qualitatively demonstrate the ability of the unconditional generator trained through Guided-GAN in producing diverse and realistic sequences. To this end, we visualise the data spaces for both the \textit{real} and \textit{generated} datasets in 2D using t-SNE \cite{tsne} in Fig. \ref{nips:mnist_space} and Fig. \ref{nips:uci_space} for \texttt{Sequential MNIST} and \texttt{UCI HAR} respectively. 

We can observe that the \textit{generated} data distribution closely follows the \textit{real} data distribution, as indicated by the dense overlap between their corresponding sequence samples. In addition, we observe a smooth interpolation in the space between different categories for the generated sequences; \textit{e.g.}, the generated samples interconnecting the \texttt{Lying} and \texttt{Sitting} activity categories in Fig. \ref{nips:uci_space}-b. Further, we depict a set of generated sequences where visual inspection of generated data for \texttt{Sequential MNIST} clearly demonstrates a conformance to the class label semantics.

\subsection{Faithfulness of Reconstructions}\label{sup_faith}

It has been reported in \cite{ali,faae} that the reconstructions of data with Bidirectional GAN (BiGAN) \cite{advfeature} and Adversarially Learned Inference (ALI) \cite{ali} are not always faithful reproductions of the inputs; \textit{in extreme cases deviating entirely from the semantic labels.} 

We conduct a set of experiments to quantitatively measure the veracity of the sequential multi-modal data reconstructions by our Guided-GAN. To this end, we rigorously explore the downstream classification tasks on \texttt{Sequential MNIST} and \texttt{UCI HAR} while considering different \textit{development datasets} (denoting the dataset used to train the supervised classifier) and \textit{evaluation datasets} (denoting the dataset used for evaluation):

\begin{itemize}
	\item \textbf{Train}---The standard training split data and labels.
	\item \textbf{Test}---The standard testing split data and labels. 
	\item \textbf{Reconstructed Train}---The reconstructed training data attained by applying $\mat{x}_{\textrm{rec}}=\mathcal{G}_{\phi}(\mathcal{E}_{\theta}(\mat{x}))$ for every sequence in the original \textbf{Train} split whilst retaining the original labels. 
	\item \textbf{Reconstructed Test}---The reconstructed test data attained by applying $\mat{x}_{\textrm{rec}}=\mathcal{G}_{\phi}(\mathcal{E}_{\theta}(\mat{x}))$ for every sequence in the original \textbf{Test} split  whilst maintaining the original labels.  
\end{itemize}

\begin{figure*}[t]%
	\centering
	\subfloat{\includegraphics[width=1.0\textwidth]{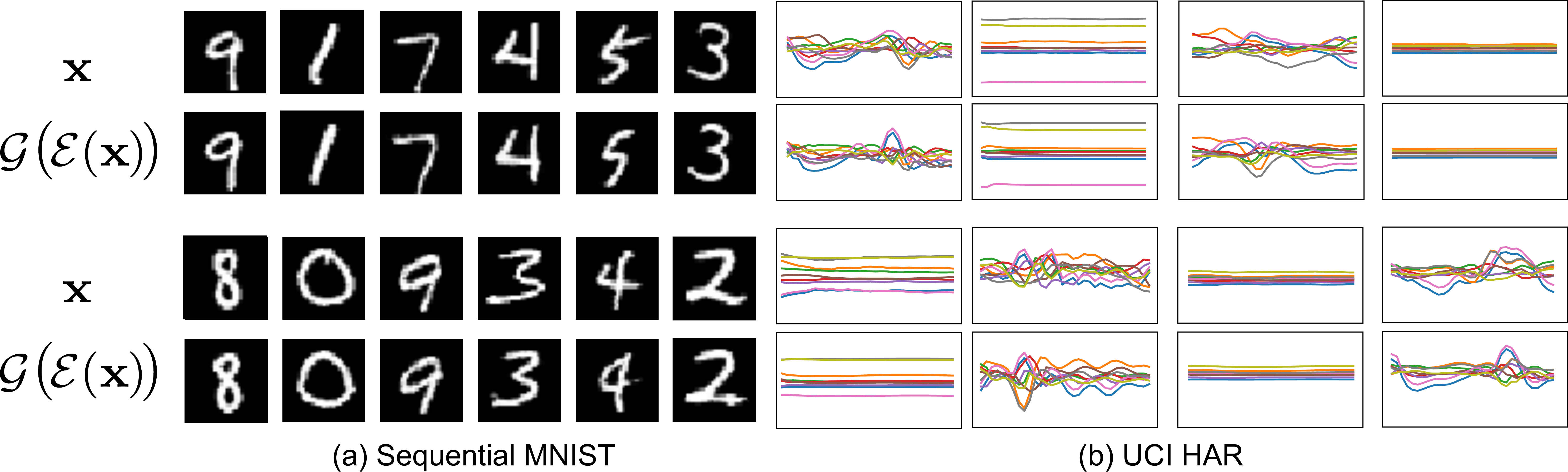}}%
	\caption[Qualitative assessment of reconstruction faithfulness]{We present qualitative results for data reconstructions with Guided-GAN for (a) \texttt{Sequential MNIST} and (b) \texttt{UCI HAR} datasets, where the odd rows represents the original data $\mat{x}$ and the even rows are the corresponding reconstructions $\mat{x}_{\textrm{rec}}=\mathcal{G}_{\phi}(\mathcal{E}_{\theta}(\mat{x}))$. Interestingly, we observe reconstructions for UCI HAR sequences reflecting different phases and variations of activity sequences whilst preserving the semantics of the reconstructed sequences.}
	\label{nips:qualitative}
\end{figure*}
\begin{table}[t]
	\caption{We quantitatively assess the faithfulness of data reconstructions through rigorous evaluation on downstream classification tasks. We report classification accuracy together with class-averaged F-score (value in parenthesis) on the holdout evaluation datasets.}\label{nips:recon}
	\centering

	\footnotesize		
	\def\arraystretch{1}
\begin{tabular}{lc|c|c}
\begin{tabular}[c]{@{}l@{}}Development \\Dataset\end{tabular} & \begin{tabular}[c]{@{}c@{}}Evaluation \\Dataset\end{tabular} & \begin{tabular}[c]{@{}c@{}}Sequential \\MNIST\end{tabular} & UCI HAR      \\ 
\toprule
Train                                                         & Test                                                         & 97.3 (97.3)                                                & 89.0 (88.9)  \\ 
\toprule
Reconstructed Train                                           & Test                                                         & 95.1 (95.0)                                                & 84.2 (84.0)  \\
Train                                                         & Reconstructed Test                                           & 93.9 (93.8)                                                & 83.4 (83.1)  \\
Reconstructed Train                                           & Reconstructed Test                                           & 94.6 (94.6)                                                & 84.9 (84.7)\\
\toprule
\end{tabular}
\end{table}

We summarise the corresponding classification performances in Table \ref{nips:recon}. From the results, across both datasets, we observe that Guided-GAN demonstrates reconstructions of reasonable faithfulness to their original semantic categories; \textit{i.e.}, substituting the original data splits---e.g. \textbf{Train}---with their corresponding reconstructions---\textbf{Reconstructed Train}---still allows learning a classifier with comparable performance to the original data splits. We further substantiate this by including qualitative samples of data reconstructions in Fig. \ref{nips:qualitative} for reference. 

\section{Related Work}\label{related-work}

\textit{Unsupervised Learning in HAR.~}A thread of studies in ubiquitous computing leverage unsupervised learning as a means for \textit{pre-training} deep activity recognition models prior to supervised fine-tuning with labeled data \cite{pd,alsheikh,autoset,HarishPlotzSelfSup2020,TangSelfHAR2021}. Exploiting the easily accessible unlabeled activity datasets in the pre-training stage guides the following discriminative training phase to better generalization performance on unseen wearable data. In this regard, \cite{alsheikh} pre-trains weights of a deep belief network using unlabeled spectrograms extracted from acceleration measurements. In another work, \cite{autoset} replaces the layer-wise pretraining with end-to-end optimization of a convolutional autoencoder applied on raw multi-modal sensor data; while \cite{HarishPlotzSelfSup2020} explores a self-supervised pre-training objective. Yet, other studies explore semi-supervised methods~\cite{TangSelfHAR2021}. However, these studies involve a subsequent supervised fine-tuning of learned parameters and thus, no isolated analysis is presented towards assessing the quality of unsupervised activity features. 

Another line of research explores \textit{unsupervised representation learning} of human activities through sensor data captured by wearables. Early efforts in this area witnessed adoption of Restricted Boltzmann Machines (RBMs) \cite{ff1} and vanilla autoencoders \cite{ff3,ff2} comprised of deep feed-forward neural networks. These approaches lack recurrent structures and thus, inherently fail to capture the temporal aspect of sensory data. Recently, \cite{role} proposed recurrent autoencoders to account for the temporal dependencies among sequential sensor samples. A LSTM encoder first processes the input multi-channel sequence into a compact latent representation. The extracted representation is then sequentially manipulated by a LSTM decoder to reproduce the sensor data. As training progresses, the encoder is expected to capture salient features required for satisfying the reconstruction task. In another recent work, \cite{motion2vec} designed a recurrent variational auto-encoder, namely Motion2Vector, by enforcing the latent representations of the encoder to follow the standard multivariate Gaussian. The effectiveness of the proposed unsupervised deep learning model is studied on  epileptic patient activity data.

\textit{Unsupervised Representation Learning with Generative Adversarial Networks.~}
GANs have demonstrated great success in approximating arbitrary complex data distributions and thus, have emerged as the state-of-the-art for realistic data generation on variety of benchmarks \cite{denton,dcgan,biggan}. Beyond data generation, recent interests investigate GAN's latent space for unsupervised feature learning through the addition of an encoder network. In \cite{infogan}, mutual information maximization is adopted to infer and gain control over a subset of latent features. In order to achieve full inference on the latents in \cite{icgan,faae}, the generator output is directly fed to an encoder network that is trained to reconstruct the latents entirely. Taking an alternative approach in learning the generator's inverse mappings, the encoder and the generator do not directly communicate in \cite{advfeature,ali}. However, the discriminator receives pairs of data and latents to conduct the discrimination task in the joint space. The resulting adversarially learned representations demonstrate state-of-the-art performance when transferred for auxiliary supervised discrimination tasks on natural images.

\textit{Generative Adversarial Networks for Multi-modal Sequential Data.~}Interest in the adoption of GANs for sequential multi-modal data has predominantly focused on realistic sequence generation. In a preliminary work, \cite{crnngan} proposes a generative adversarial model that operates on continuous sequential data and applies the framework on a collection of classical music. The resulting GAN adopts LSTM networks for generator and discriminator to generate polyphonic music. Applying architectural modifications, the authors in \cite{medical} develop a recurrent GAN to produce synthetic medical time-series data. The approach is further extended to exploit data annotations for conditional generation in order to substitute sensitive patient records. Particularly for HAR applications, authors in \cite{sensegen} employ GANs to generate synthetic sensor data preserving statistics of smartphone accelerometer sensor traces. Similarly in \cite{sensorygans}, authors attempt to synthesize sensory data captured by wearable sensors for human activity recognition. Due to the challenging nature and heterogeneity of sequential sensor data, the authors faced difficulties developing a unified GAN to cover the multi-modal distribution of human actions. Accordingly, the framework leverages data annotations to train independent activity-specific GANs and achieve stable training. 

In another work, \cite{wifi} trains a GAN with pre-processed WiFi signals and generates new patterns with similar statistical features to expand the available training data. The augmented dataset is subsequently used to train a LSTM recognition model to infer human actions in indoor environments. Both discussed studies operate on pre-processed input features instead of raw multi-channel sensor data. Besides synthetic sequence generation, recent semi-supervised activity recognition has also benefited from adversarial training. In this regard, \cite{semisup} assumes availability of limited labeled data during training and develops a convolutional adversarial autoencoder to encode multi-channel sequences into discrete class representations as well as continuous latent variables. The resulting framework is shown to achieve additional performance boost by exploiting unlabeled data. Similarly in a semi-supervised context, \cite{transfer} investigates the problem of cross-subject adversarial knowledge transfer using GANs in the domain of human activity recognition. 

\vspace{2mm}
\noindent\textbf{Summary.~}In the realm of unsupervised learning for HAR applications, training variants of autoencoders with element-wise reconstruction objective constitutes the dominant unsupervised representation learning pipeline. While in the visual domain, deep generative models have demonstrated great potential in unsupervised learning of enriched features for natural images, their application for human activity recognition with sequential sensor data is mainly limited to synthesizing artificial sequences that resemble wearable sensor sequences. Accordingly, this paper presents a first rigorous study of GAN-based alternatives for unsupervised learning of activity representations in our attempts to bridge the gap between unsupervised representation learning and generative adversarial networks for human activity recognition with wearables.
\begin{figure*}[t]%
	\centering
	\subfloat{\includegraphics[width=1\textwidth]{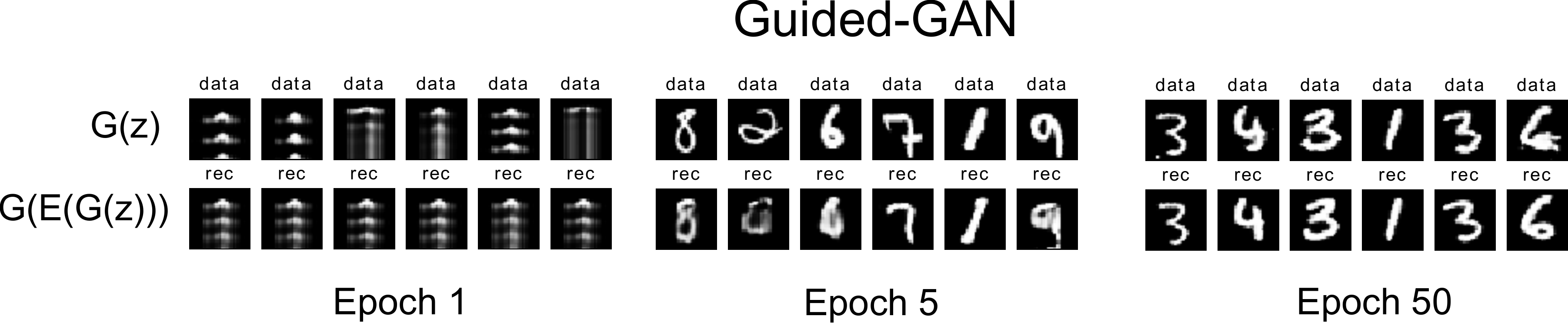}}\\
	\subfloat{\includegraphics[width=1\textwidth]{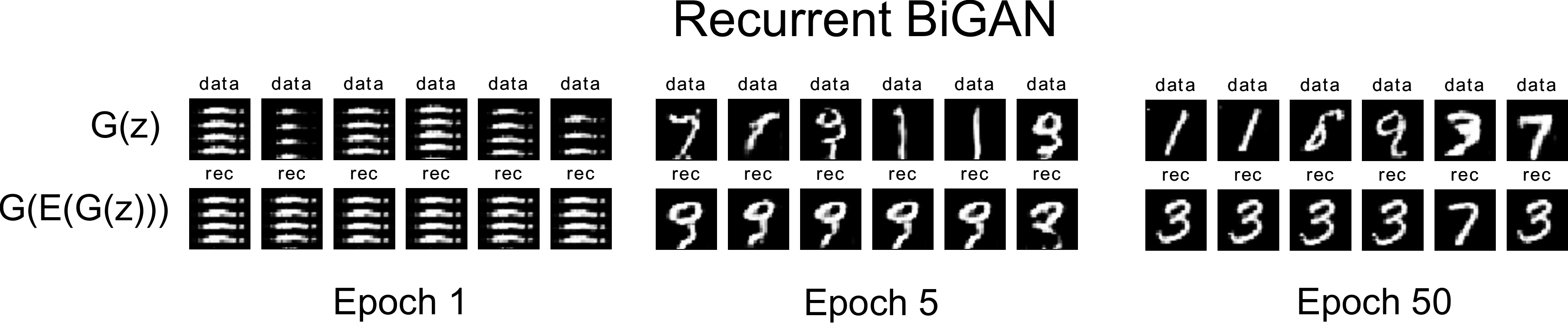}} 
	\caption{Training comparisons between our proposed Guided-GAN and \textit{recurrent} BiGAN. As illustrated, we observed through multiple experiments, the inability of the recurrent BiGAN to uncover the generator's inverse mapping.}
	\label{bigan-exp-train}
\end{figure*}

\begin{figure}[h]%
	\centering
	\subfloat{\includegraphics[width=0.4\textwidth]{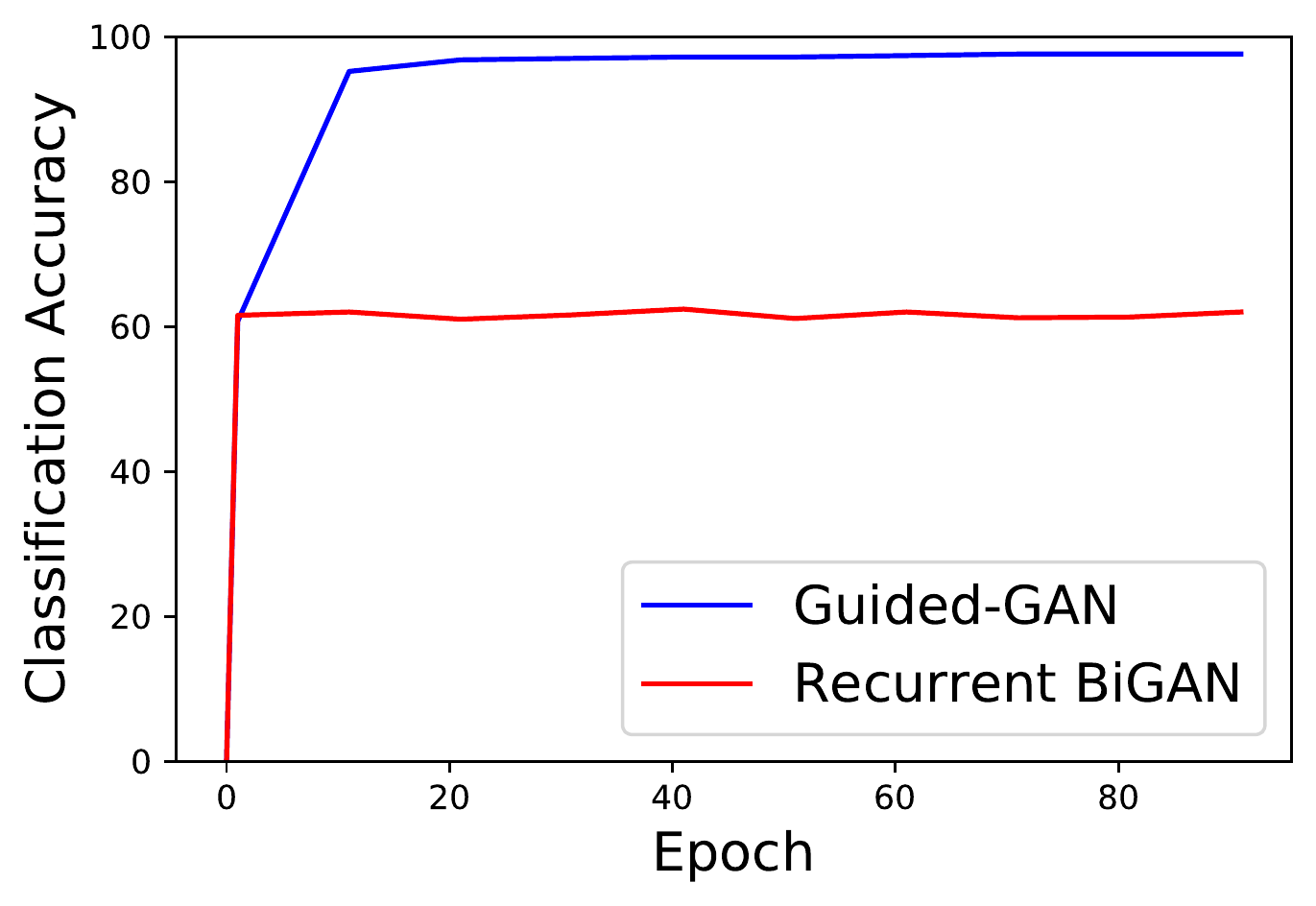}}%
	\caption{A comparison of classification accuracy achieved on holdout test split for Guided-GAN and recurrent BiGAN evaluated at set epocs during training process.}
	\label{bigan-cmp-plot}
\end{figure}

\section{Conclusions}
We propose \textit{Guided-GAN} in our efforts to abridge unsupervised representation learning and generative adversarial networks for advancing the field of unsupervised learning for wearable-HAR. Recognizing the inherent temporal dependencies within the captured sequential data, we design a new recurrent network architecture incorporating in a bidirectional GAN framework. We share the key insight that the discriminator adversarial feedback alone may be \textit{insufficient} to uncover the generator's inverse mapping. Hence, Guided-GAN leverages an intuitive formulation to alleviate the burden on the discriminator in achieving inverting generators and encoders. When evaluated on three downstream sequence classification benchmarks, our learned sequence representations outperform existing unsupervised approaches whilst closely approaching the performance of fully supervised learned features.

\section{Appendix}\label{apd:biganexp}

We present our results from experimental evaluations demonstrating the difficulty in convergence observed in training a \textit{recurrent} BiGAN~\cite{biggan}.

\vspace{1mm}
\noindent\textbf{Comparing Training of Guided-GAN with Recurrent BiGAN.~} As mentioned in Section~\ref{sec:our-framework}, our attempts to train recurrent BiGANs without our proposed manifold distance minimization terms were unsuccessful. For reference, we present empirical results obtained from training recurrent BiGAN as well as our Guided-GAN on \texttt{Sequential MNIST} for ease of visual inspections. 

To observe the behavior of generators and encoders, we visualize randomly generated samples $\hat{\mat{x}}=\mathcal{G}_{\phi}(\mat{z})$ at different stages of the training process together with their corresponding reconstructions $\mat{x}_{\textrm{rec}}=\mathcal{G}_{\phi}(\mathcal{E}_{\theta}(\hat{\mat{x}}))$ for both \textit{recurrent} BiGAN and our proposed Guided-GAN in Fig.~\ref{bigan-exp-train}. In particular, we observed that the sole discriminator in BiGAN was not able to guide the recurrent encoder towards uncovering the generator's inverse mapping function. Thus, no useful representations were obtained and accordingly, extremely low downstream classification performance was achieved. In contrast, our proposed Guided-GAN successfully inverts the generator and encoder at the very early stages of training process. We further provide the downstream classification performance achieved by both approaches in Fig.~\ref{bigan-cmp-plot} at set epochs in the training process.

\bibliographystyle{IEEEtran}
\bibliography{referencelist}

\begin{thebibliography}{10}
\providecommand{\url}[1]{#1}
\csname url@samestyle\endcsname
\providecommand{\newblock}{\relax}
\providecommand{\bibinfo}[2]{#2}
\providecommand{\BIBentrySTDinterwordspacing}{\spaceskip=0pt\relax}
\providecommand{\BIBentryALTinterwordstretchfactor}{4}
\providecommand{\BIBentryALTinterwordspacing}{\spaceskip=\fontdimen2\font plus
\BIBentryALTinterwordstretchfactor\fontdimen3\font minus
  \fontdimen4\font\relax}
\providecommand{\BIBforeignlanguage}[2]{{%
\expandafter\ifx\csname l@#1\endcsname\relax
\typeout{** WARNING: IEEEtran.bst: No hyphenation pattern has been}%
\typeout{** loaded for the language `#1'. Using the pattern for}%
\typeout{** the default language instead.}%
\else
\language=\csname l@#1\endcsname
\fi
#2}}
\providecommand{\BIBdecl}{\relax}
\BIBdecl

\bibitem{mannini2017activity}
A.~Mannini, M.~Rosenberger, W.~L. Haskell, A.~M. Sabatini, and S.~S. Intille,
  ``Activity recognition in youth using single accelerometer placed at wrist or
  ankle,'' \emph{Medicine and Science in Sports and Exercise}, vol.~49, no.~4,
  p. 801, 2017.

\bibitem{robertoFallsPO2017}
R.~L. Shinmoto~Torres, R.~Visvanathan, D.~Abbott, K.~D. Hill, and D.~C.
  Ranasinghe, ``A battery-less and wireless wearable sensor system for
  identifying bed and chair exits in a pilot trial in hospitalized older
  people,'' \emph{PLOS ONE}, vol.~12, no.~10, pp. 1--25, 10 2017.

\bibitem{Govercin2010UserReqFallDetect}
M.~G{\"o}vercin, Y.~K{\"o}ltzsch, M.~Meis, S.~Wegel, M.~Gietzelt, J.~Spehr,
  S.~Winkelbach, M.~Marschollek, and E.~Steinhagen-Thiessen, ``Defining the
  user requirements for wearable and optical fall prediction and fall detection
  devices for home use,'' \emph{Informatics for Health and Social Care},
  vol.~35, no. 3-4, pp. 177--187, 2010.

\bibitem{healthapp}
J.~Frank, S.~Mannor, and D.~Precup, ``Activity and gait recognition with
  time-delay embeddings,'' in \emph{AAAI Conference on Artificial Intelligence
  (AAAI)}, 2010.

\bibitem{pd}
N.~Y. Hammerla, J.~Fisher, P.~Andras, L.~Rochester, R.~Walker, and
  T.~Pl{\"o}tz, ``{PD} disease state assessment in naturalistic environments
  using deep learning,'' in \emph{AAAI Conference on Artificial Intelligence
  (AAAI)}, 2015.

\bibitem{asangiweariswc}
A.~Jayatilaka, Q.~H. Dang, S.~J. Chen, R.~Visvanathan, C.~Fumeaux, and D.~C.
  Ranasinghe, ``Designing batteryless wearables for hospitalized older
  people,'' in \emph{Proceedings of the International Symposium on Wearable
  Computers (ISWC)}, 2019, p. 91–95.

\bibitem{health4}
A.~{Fellger}, G.~{Sprint}, D.~{Weeks}, E.~{Crooks}, and D.~J. {Cook},
  ``Wearable device-independent next day activity and next night sleep
  prediction for rehabilitation populations,'' \emph{IEEE J. Transl. Eng.
  Health Med.}, vol.~8, pp. 1--9, 2020.

\bibitem{chesser2018bedexit}
M.~{Chesser}, A.~{Jayatilaka}, R.~{Visvanathan}, C.~{Fumeaux}, A.~{Sample}, and
  D.~C. {Ranasinghe}, ``Super low resolution {RF} powered accelerometers for
  alerting on hospitalized patient bed exits,'' in \emph{IEEE International
  Conference on Pervasive Computing and Communications (PerCom)}, 2019, pp.
  1--10.

\bibitem{role}
H.~Haresamudram, D.~V. Anderson, and T.~Pl{\"o}tz, ``On the role of features in
  human activity recognition,'' in \emph{Proceedings of the International
  Symposium on Wearable Computers (ISWC)}, 2019, pp. 78--88.

\bibitem{motion2vec}
L.~Bai, C.~Yeung, C.~Efstratiou, and M.~Chikomo, ``{Motion2Vector}:
  Unsupervised learning in human activity recognition using wrist-sensing
  data,'' in \emph{Proceedings of the ACM International Symposium on Wearable
  Computers (ISWC)}, 2019, p. 537–542.

\bibitem{gan}
I.~Goodfellow, J.~Pouget-Abadie, M.~Mirza, B.~Xu, D.~Warde-Farley, S.~Ozair,
  A.~Courville, and Y.~Bengio, ``Generative adversarial nets,'' in
  \emph{Conference on Neural Information Processing Systems (NIPS)}, 2014, pp.
  2672--2680.

\bibitem{advfeature}
J.~Donahue, P.~Kr{\"a}henb{\"u}hl, and T.~Darrell, ``Adversarial feature
  learning,'' in \emph{The International Conference on Learning Representations
  (ICLR)}, 2017.

\bibitem{ali}
V.~Dumoulin, I.~Belghazi, B.~Poole, O.~Mastropietro, A.~Lamb, M.~Arjovsky, and
  A.~Courville, ``Adversarially learned inference,'' in \emph{The International
  Conference on Learning Representations (ICLR)}, 2017.

\bibitem{icgan}
G.~Perarnau, J.~van~de Weijer, B.~Raducanu, and J.~M. \'Alvarez, ``{Invertible
  Conditional GANs for image editing},'' in \emph{NIPS Workshop on Adversarial
  Training}, 2016.

\bibitem{infogan}
X.~Chen, Y.~Duan, R.~Houthooft, J.~Schulman, I.~Sutskever, and P.~Abbeel,
  ``{InfoGAN}: Interpretable representation learning by information maximizing
  generative adversarial nets,'' in \emph{Conference on Neural Information
  Processing Systems (NIPS)}, 2016, p. 2180–2188.

\bibitem{faae}
J.~Zhang, H.~Dang, H.~K. Lee, and E.-C. Chang, ``Flipped-adversarial
  autoencoders,'' \emph{arXiv preprint arXiv:1802.04504}, 2018.

\bibitem{bigbigan}
J.~Donahue and K.~Simonyan, ``Large scale adversarial representation
  learning,'' in \emph{Conference on Neural Information Processing Systems
  (NIPS)}, 2019, pp. 10\,542--10\,552.

\bibitem{crnngan}
O.~Mogren, ``{C-RNN-GAN}: Continuous recurrent neural networks with adversarial
  training,'' \emph{arXiv preprint arXiv:1611.09904}, 2016.

\bibitem{medical}
C.~Esteban, S.~L. Hyland, and G.~R{\"a}tsch, ``Real-valued (medical) time
  series generation with recurrent conditional gans,'' \emph{arXiv preprint
  arXiv:1706.02633}, 2017.

\bibitem{sensegen}
M.~Alzantot, S.~Chakraborty, and M.~B. Srivastava, ``{SenseGen}: A deep
  learning architecture for synthetic sensor data generation,'' in \emph{IEEE
  International Conference on Pervasive Computing and Communications Workshops
  (PerCom Workshops)}, 2017.

\bibitem{sensorygans}
J.~Wang, Y.~Chen, Y.~Gu, Y.~Xiao, and H.~Pan, ``{SensoryGANs}: an effective
  generative adversarial framework for sensor-based human activity
  recognition,'' in \emph{International Joint Conference on Neural Networks
  (IJCNN)}, 2018, pp. 1--8.

\bibitem{wifi}
P.~F. Moshiri, H.~Navidan, R.~Shahbazian, S.~A. Ghorashi, and D.~Windridge,
  ``Using {GAN} to enhance the accuracy of indoor human activity recognition,''
  \emph{arXiv preprint arXiv:2004.11228}, 2020.

\bibitem{timegan}
J.~Yoon, D.~Jarrett, and M.~van~der Schaar, ``Time-series generative
  adversarial networks,'' in \emph{Conference on Neural Information Processing
  Systems (NIPS)}, 2019, pp. 5508--5518.

\bibitem{dcgan}
A.~Radford, L.~Metz, and S.~Chintala, ``Unsupervised representation learning
  with deep convolutional generative adversarial networks,'' \emph{arXiv
  preprint arXiv:1511.06434}, 2015.

\bibitem{miyato}
T.~Miyato, T.~Kataoka, M.~Koyama, and Y.~Yoshida, ``Spectral normalization for
  generative adversarial networks,'' in \emph{The International Conference on
  Learning Representations (ICLR)}, 2018.

\bibitem{salimans}
T.~Salimans, I.~Goodfellow, W.~Zaremba, V.~Cheung, A.~Radford, and X.~Chen,
  ``Improved techniques for training gans,'' in \emph{Conference on Neural
  Information Processing Systems (NIPS)}, 2016, pp. 2234--2242.

\bibitem{karras}
T.~Karras, T.~Aila, S.~Laine, and J.~Lehtinen, ``Progressive growing of gans
  for improved quality, stability, and variation,'' \emph{The International
  Conference on Learning Representations (ICLR)}, 2018.

\bibitem{audeep}
M.~Freitag, S.~Amiriparian, S.~Pugachevskiy, N.~Cummins, and B.~Schuller,
  ``audeep: Unsupervised learning of representations from audio with deep
  recurrent neural networks,'' \emph{Journal of Machine Learning Research},
  vol.~18, no. 173, pp. 1--5, 2018.

\bibitem{modegan}
T.~Che, Y.~Li, A.~P. Jacob, Y.~Bengio, and W.~Li, ``Mode regularized generative
  adversarial networks,'' in \emph{The International Conference on Learning
  Representations (ICLR)}, 2017.

\bibitem{uci}
D.~Anguita, A.~Ghio, L.~Oneto, X.~Parra, and J.~L. Reyes-Ortiz, ``A public
  domain dataset for human activity recognition using smartphones.'' in
  \emph{European Symposium on Artificial Neural Networks (ESANN)}, 2013.

\bibitem{usc}
M.~Zhang and A.~A. Sawchuk, ``Usc-had: a daily activity dataset for ubiquitous
  activity recognition using wearable sensors,'' in \emph{Proceedings of the
  ACM Conference on Ubiquitous Computing (UbiComp)}, 2012, pp. 1036--1043.

\bibitem{mnist}
Y.~LeCun, L.~Bottou, Y.~Bengio, and P.~Haffner, ``Gradient-based learning
  applied to document recognition,'' \emph{Proceedings of the IEEE}, vol.~86,
  no.~11, pp. 2278--2324, 1998.

\bibitem{eval}
R.~Zhang, P.~Isola, and A.~A. Efros, ``Colorful image colorization,'' in
  \emph{European Conference on Computer Vision (ECCV)}, 2016, pp. 649--666.

\bibitem{adam}
D.~P. Kingma and J.~Ba, ``Adam: A method for stochastic optimization,'' in
  \emph{The International Conference on Learning Representations (ICLR)}, 2015.

\bibitem{tsne}
L.~v.~d. Maaten and G.~Hinton, ``Visualizing data using t-sne,'' in
  \emph{{Journal of Machine Learning Research}}, 2008.

\bibitem{alsheikh}
M.~A. Alsheikh, A.~Selim, D.~Niyato, L.~Doyle, S.~Lin, and H.-P. Tan, ``Deep
  activity recognition models with triaxial accelerometers,'' in
  \emph{Workshops at the AAAI Conference on Artificial Intelligence}, 2016.

\bibitem{autoset}
A.~A. Varamin, E.~Abbasnejad, Q.~Shi, D.~C. Ranasinghe, and H.~Rezatofighi,
  ``Deep auto-set: A deep auto-encoder-set network for activity recognition
  using wearables,'' in \emph{Proceedings of the EAI International Conference
  on Mobile and Ubiquitous Systems: Computing, Networking and Services
  (MobiQuitous)}, 2018, p. 246–253.

\bibitem{HarishPlotzSelfSup2020}
H.~Haresamudram, A.~Beedu, V.~Agrawal, P.~L. Grady, I.~Essa, J.~Hoffman, and
  T.~Pl\"{o}tz, ``Masked reconstruction based self-supervision for human
  activity recognition,'' in \emph{Proceedings of the International Symposium
  on Wearable Computers (ISWC)}, 2020, p. 45–49.

\bibitem{TangSelfHAR2021}
C.~I. Tang, I.~Perez-Pozuelo, D.~Spathis, S.~Brage, N.~Wareham, and C.~Mascolo,
  ``{SelfHAR}: Improving human activity recognition through self-training with
  unlabeled data,'' \emph{Proc. ACM Interact. Mob. Wearable Ubiquitous
  Technol.}, vol.~5, no.~1, 2021.

\bibitem{ff1}
B.~Almaslukh, J.~AlMuhtadi, and A.~Artoli, ``An effective deep autoencoder
  approach for online smartphone-based human activity recognition,'' \emph{Int.
  J. Comput. Sci. Netw. Secur}, vol.~17, no.~4, pp. 160--165, 2017.

\bibitem{ff3}
T.~Pl{\"o}tz, N.~Y. Hammerla, and P.~L. Olivier, ``Feature learning for
  activity recognition in ubiquitous computing,'' in \emph{International Joint
  Conference on Artificial Intelligence (IJCAI)}, 2011.

\bibitem{ff2}
J.~Wang, X.~Zhang, Q.~Gao, H.~Yue, and H.~Wang, ``Device-free wireless
  localization and activity recognition: A deep learning approach,'' \emph{IEEE
  Trans. Veh. Technol.}, vol.~66, no.~7, pp. 6258--6267, 2016.

\bibitem{denton}
E.~Denton, S.~Chintala, A.~Szlam, and R.~Fergus, ``Deep generative image models
  using a laplacian pyramid of adversarial networks,'' in \emph{Conference on
  Neural Information Processing Systems (NIPS)}, 2015, p. 1486–1494.

\bibitem{biggan}
A.~Brock, J.~Donahue, and K.~Simonyan, ``Large scale {GAN} training for high
  fidelity natural image synthesis,'' in \emph{The International Conference on
  Learning Representations (ICLR)}, 2019.

\bibitem{semisup}
D.~Balabka, ``Semi-supervised learning for human activity recognition using
  adversarial autoencoders,'' in \emph{Adjunct Proceedings of the ACM
  International Joint Conference on Pervasive and Ubiquitous Computing and
  Proceedings of the ACM International Symposium on Wearable Computers (ISWC)},
  pp. 685–--688.

\bibitem{transfer}
E.~Soleimani and E.~Nazerfard, ``Cross-subject transfer learning in human
  activity recognition systems using generative adversarial networks,''
  \emph{arXiv preprint arXiv:1903.12489}, 2019.

\end{thebibliography}

\end{document}